\documentclass[letterpaper, 10 pt, conference]{ieeeconf}  %

\usepackage{tabularx}
\usepackage{threeparttable}
\usepackage{graphicx}
\usepackage{xcolor}
\usepackage[export]{adjustbox}
\usepackage{listliketab}
\usepackage{tabulary}

\usepackage{booktabs}
\usepackage{array}
\usepackage{newtxtext}
\usepackage{newtxmath}
\let\labelindent\relax
\usepackage{enumitem}
\usepackage{soul}

\graphicspath{ {./images/} }

\IEEEoverridecommandlockouts                              %

\title{\LARGE \bf
\centering Evaluating Visual Odometry Methods for Autonomous Driving in Rain
}
\author{Yu Xiang Tan, Marcel Bartholomeus Prasetyo, Mohammad Alif Daffa, \\ Deshpande Sunny Nitin and Malika Meghjani
    \thanks{Yu Xiang Tan, Marcel Bartholomeus Prasetyo, Mohammad Alif Daffa, Deshpande Sunny Nitin and Malika Meghjani are with Singapore University of Technology and Design, Singapore. {\tt\small \{yuxiang\_tan, mohammad\_alif, sunny\_deshpande\}@mymail.sutd.edu.sg, \{marcel\_prasetyo, malika\_meghjani\}@sutd.edu.sg}} %
}

\date{March 2023}

\begin{document}

\maketitle
\thispagestyle{empty}
\pagestyle{empty}

\begin{abstract}

The increasing demand for autonomous vehicles has created a need for robust navigation systems that can also operate effectively in adverse weather conditions. Visual odometry is a technique used in these navigation systems, enabling the estimation of vehicle position and motion using input from onboard cameras. However, visual odometry accuracy can be significantly impacted in challenging weather conditions, such as heavy rain, snow, or fog. In this paper, we evaluate a range of visual odometry methods, including our DROID-SLAM based heuristic approach. Specifically, these algorithms are tested on both clear and rainy weather urban driving data to evaluate their robustness. We compiled a dataset comprising of a range of rainy weather conditions from different cities. This includes, the Oxford Robotcar dataset from Oxford, the 4Seasons dataset from Munich and an internal dataset collected in Singapore. We evaluated different visual odometry algorithms for both monocular and stereo camera setups using the Absolute Trajectory Error (ATE). From the range of approaches evaluated, our findings suggest that the Depth and Flow for Visual Odometry (DF-VO) algorithm with monocular setup performed the best for short range distances ($ < 500m$) and our proposed DROID-SLAM based heuristic approach for the stereo setup performed relatively well for long-term localization. 
Both VO algorithms suggested a need for a more robust sensor fusion based approach for localization in rain.

\end{abstract}

\section{INTRODUCTION}

\begin{figure}[ht]
    \centering
    \newcolumntype{C}{>{\centering\arraybackslash}X}
    \begin{tabularx}{\columnwidth}{CC}
    \fontfamily{cmr}\selectfont
        \includegraphics[width=0.48\columnwidth]{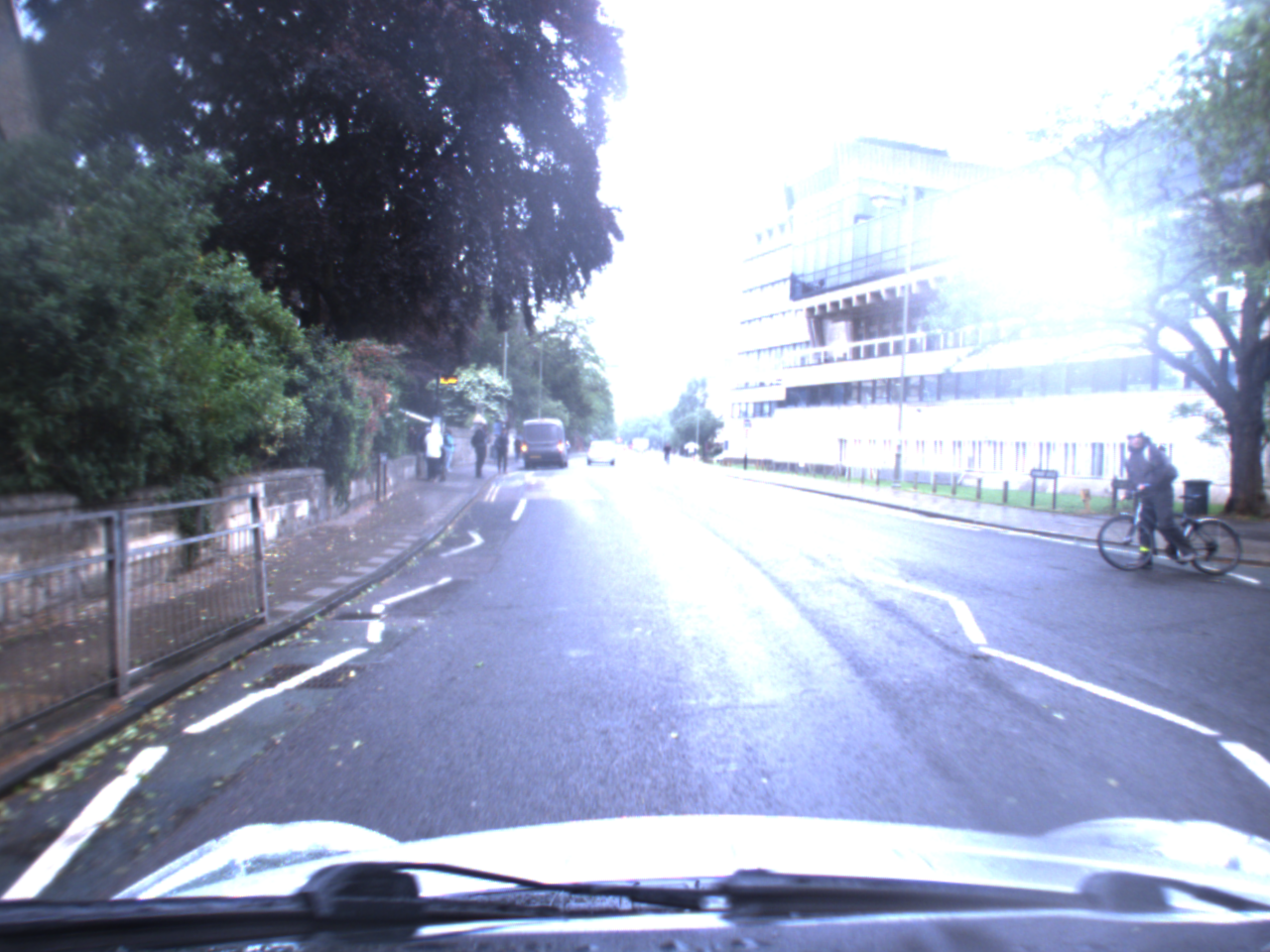} &
        \includegraphics[width=0.48\columnwidth]{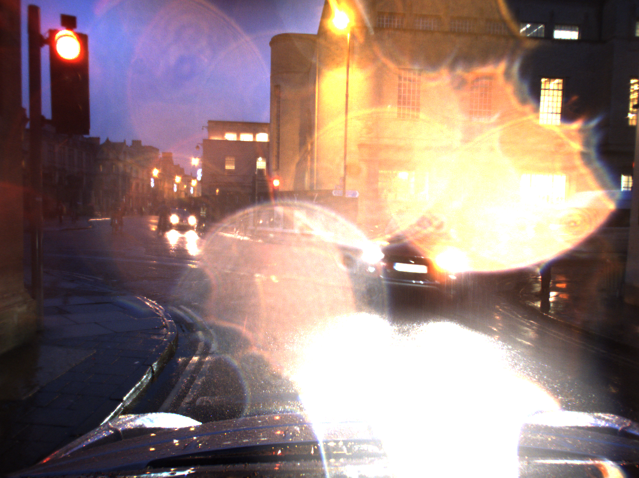}\\
        \scriptsize{Oxford Robotcar (Day)} & \scriptsize{Oxford Robotcar (Night)} \\
        \includegraphics[height=.3\columnwidth,width=0.48\columnwidth]{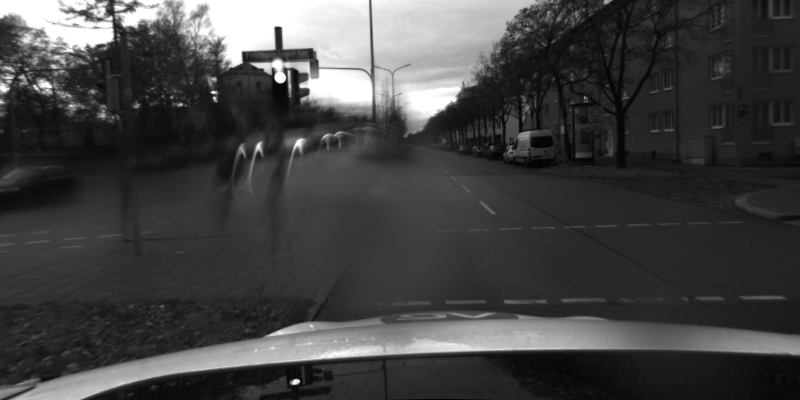} &
        \includegraphics[height=.3\columnwidth,width=0.48\columnwidth]{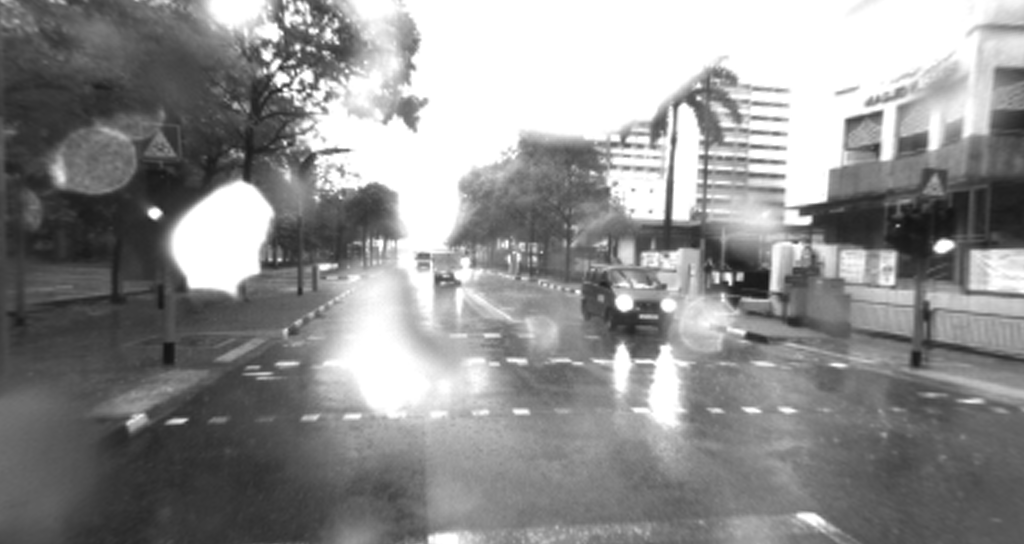}  \\
        \scriptsize{4Seasons} & \scriptsize{Singapore} \\
    \end{tabularx}
    \caption{Sample images from the three datasets.}
    \label{fig:datasets}
\vspace{-0.02\textheight}    
\end{figure}

Visual Odometry (VO) is a cost-effective localization solution for autonomous urban driving. However, visual data can be easily compromised in adverse weather conditions such as rain, fog or snow. In rain, images are occluded by raindrops on the camera lenses and rain streaks reduce the visibly of the background objects \cite{garg_vision_2007}. Lens flare also appear due to rain which further reduces the visibility of the scene \cite{ulfwi_adherent_nodate} as shown in Fig. \ref{fig:datasets}. These adverse weather effects could negatively impact visual odometry algorithms designed and trained on clear weather conditions \cite{bahnsen_rain_2019}. This calls for a robust localization algorithm to enable autonomous vehicles to operate in all-weather conditions. 

In this paper, we evaluate across a range of VO algorithms, including our DROID-SLAM based heuristic approach \cite{tan_localization_2023}, for urban driving in rain. Our aim is to identify the VO algorithm that performs relatively well for robust localization in rainy weather. We compiled the available open-source rain datasets and augmented them with our internal rain dataset to create a comprehensive suite of datasets for evaluation. The open-source datasets comprise of the Oxford Robotcar \cite{maddern_1_2017} and the 4Seasons datasets \cite{wenzel_4seasons_2020}. Our internal dataset was collected in Singapore while Oxford Robotcar and 4Seasons datasets are from Oxford and Munich respectively. Thus, the resultant combined dataset used in our study contains a wide range of road and rain conditions from climatically and geographically different cities. Sample images from the three datasets are presented in Fig. \ref{fig:datasets}.

We analyze the strengths and limitations of various approaches and provide insights into future research directions that can improve the robustness and reliability of visual odometry based algorithms in these scenarios. Our contributions are: (a) a comprehensive evaluation of existing VO algorithms for both clear and rain conditions and (b) an analysis of strengths and limitations of different VO algorithms for rain conditions.

\vspace{-0.001\textheight}

\section{RELATED WORK} \label{related_work}
There are three parts of related work for this paper: (a) robust sensors and sensor fusion based localization algorithms in adverse weather, (b) robust visual feature extraction in adverse weather scenarios and (c) visual odometry algorithms.

\subsection{Localization in Adverse Weather} 
The diverse range of noise artefacts caused by adverse weather makes localization a challenging problem. The state-of-the-art approaches for localization in adverse weather, either localize based on robust sensors or perform sensor fusion. Zhang et al. discussed the impacts of adverse weather on autonomous driving, across multiple weather conditions, and sensors \cite{zhang_autonomous_2021}. Meanwhile, our paper performs an in-depth analysis on the effects of rain conditions and evaluates a wide range of visual odometry methods on real-world rain datasets. 

\subsubsection{Utilizing Robust Sensors}
Robust sensors such as Radar and LiDAR are less affected by adverse weather conditions when compared to the vision data. Thus, recent approaches \cite{hong_radarslam_2020, ort_autonomous_2020, barnes_under_2020} adopt to utilize such sensors for adverse weather conditions. However, such sensors are more expensive compared to cameras and require higher computational requirements. 

\subsubsection{Sensor Fusion Approaches}
Sensor fusion utilizes multiple sensors together to perform localization for difficult conditions. An example would be Visual-Inertial Odometry where camera images are used together with the Inertial Measurement Unit (IMU) to improve robustness of the localization algorithm \cite{campos_orb-slam3_2021, von_stumberg_direct_2018}. Other examples include GPS-SLAM \cite{kiss-illes_gps-slam_2019} which combines GPS data with VO while Brubaker et al \cite{brubaker_map-based_2016} proposed a map-based approach which combines map data with VO. In this paper, we focus our evaluation on pure VO algorithms to find a suitable VO component for any sensor fusion approach for localization in rainy weather. 

\subsection{Vision in Adverse Weather}
Apart from localization, many other applications suffer from poor visual data due to adverse weather conditions. There are two main ideas to improve robustness of vision algorithms in adverse weather: (a) improving the robustness of visual features and (b) removing the noise caused by adverse weather.

\subsubsection{Robust Feature Detectors}
This category of approaches focuses on improving reliability of visual odometry by improving the feature detectors. Feature detectors form the basis of VO methods, where poorly detected features directly affect localization accuracy. Thus, if the feature detectors are robust, it would in turn improve the robustness of VO in adverse weather. Algorithms such as R2D2 \cite{revaud_r2d2_2019} and D2-Net \cite{dusmanu_d2-net_2019} are designed to perform accurate feature detection in presence of illumination and viewpoint changes. However, such methods are not directly trained and tested on adverse weather data.

\subsubsection{Removing Noise Artefacts}
Noise artefacts such as raindrops on the camera lens or lens flare can be removed using a neural network. Methods such as \cite{yu_visual_2021,porav_adversarial_2018} utilize pairs of adverse weather and clear weather data to train the model to remove the artefacts caused by adverse weather. This allows the algorithm to run in adverse weather the way it does in clear weather. However, it is difficult to collect pairs of adverse weather and clear weather data for a new environment and such noise removal algorithms are not proven to be generalizable enough to work properly on out-of-distribution data. In this paper, we aim to quantitatively evaluate the readiness of existing VO methods to run in adverse weather without such noise removal algorithms. 

\subsection{Visual Odometry} \label{specific_vo}
Yousif et al. \cite{yousif_overview_2015} provides an overview of the techniques involved in visual odometry and visual SLAM, while Kazerouni et al. \cite{abaspur_kazerouni_survey_2022} discusses state-of-the-art visual SLAM approaches. However, both survey papers do not analyze and evaluate in detail the capability of visual odometry in challenging scenarios. Agostinho et al. \cite{agostinho_practical_2022} evaluates visual odometry methods in challenging scenarios such as the presence of vegetation, tunnels and dynamic objects but not in adverse weather conditions. The authors highlight that visual odometry suffers in these challenging conditions due to the lack of good visual features. Similarly, our contribution lies in evaluation and analysis of visual odometry methods in challenging conditions but specific to rain conditions. In the following sections we discuss different categories of VO approaches. %

\subsubsection{Direct vs Indirect Approaches}
Direct approaches \cite{engel_direct_2016,zhan_df-vo_2021,fleet_lsd-slam_2014} minimize the photometric loss while indirect approaches \cite{campos_orb-slam3_2021,teed_droid-slam_2021,wang_tartanvo_2020,schenk_robust_2017,gomez-ojeda_robust_2016} minimize the reprojection error. Though direct approaches are resistant to photometric noise, they are computationally expensive as they solve a more complex optimization problem. Indirect approaches are more resistant to geometric noise but are more susceptible to scenes with less texture \cite{engel_direct_2016}.

\subsubsection{Dense vs Sparse Approaches}
Dense approaches use the entire image while sparse approaches use a subset of the image. Sparse methods identify keypoints \cite{engel_direct_2016} or edges \cite{gomez-ojeda_robust_2016, schenk_robust_2017} to be used in their optimization formulation. Although dense methods provide more information, the trade-off is that they have a higher computational cost. Sparse methods provide the option of having lower computational cost for less information. In rainy weather, occlusions caused by raindrops significantly reduce the amount of features identified if the raindrops land in a feature-rich region. Thus, sparse methods might delocalize due to a lack of features and dense methods might not work if they use the raindrop-occluded regions as part of their set of features. 
Our findings suggest that adopting a dense approach provides more robustness as long as the identified features are reliable. 

\subsubsection{Learning-based vs Classical Approaches}
Recent efforts use machine learning models \cite{wang_tartanvo_2020, zhan_df-vo_2021} to identify features learnt from large amounts of data. Such methods are ideal for algorithms operating within the same distribution that they are trained on and achieve localization accuracy much higher than classical methods. However, it is unknown whether a learning algorithm is able to generalize to different environments or different weather conditions. We implement learning-based localization algorithms trained on either urban driving or aerial datasets, and evaluate them on urban driving datasets in different cities and different weather conditions to test their generalizability. Classical methods use handcrafted feature detection algorithms \cite{campos_orb-slam3_2021,engel_direct_2016} which we test to compare against learning-based methods. Our findings suggest that the learning-based algorithms delocalize less compared to classical methods in rain conditions.

\subsubsection{Mixed Approaches}
Mixed approaches use both methods from the binary categories above. SVO uses both direct and indirect methods \cite{forster_svo_2014} while DF-VO uses both dense and sparse methods \cite{zhan_df-vo_2021}. CNN-SVO combines both learning-based and classical methods, where a learning-based model is used to predict depth information which is fed into the classical VO model for optimizing pose \cite{loo_cnn-svo_2018}. Although mixed approaches are designed to bring out the best of both methods, they might also include the limitations of both methods such as the assumptions made in both the direct and indirect formulations. Our findings suggest that combining dense and sparse approach perform relatively well compared to a purely dense or sparse approach in rain conditions. Also, introducing a learning-based approach for estimating depth improves the localization accuracy for rain conditions as well.

\section{EVALUATED VISUAL ODOMETRY ALGORITHMS}
\label{chosen_methods}
We chose seven open-source localization algorithms along with our proposed approach to perform evaluation in both clear and rain conditions. 
These algorithms were specifically chosen across three categories:(a) dense vs sparse, (b) classical vs learning, and (c) direct vs indirect. 
In this section, we give a brief overview of each of the algorithms implemented and highlight parts of their design that are affected by rain. Particularly, we will be considering the feature extraction and matching strategy alongside methods of tackling failures.

\subsection{Direct Sparse Odometry (DSO)} 
DSO is a classical, sparse and direct method that uses keyframes to perform a joint optimization of camera pose and 3D world model. It was tested on three datasets: TUM monoVO, EuroC MAV \cite{burri_euroc_2016} and ICL-NUIM \cite{handa_benchmark_2014} datasets and it performed robustly with accurate localization results. DSO is able to handle low textured environments using its novel feature selection strategy and has multiple methods of detecting outliers or tracking failures.

The feature selection strategy used in DSO aims to find an even spread of features across the image and at the same time select unique features in each region of the image. To create an even spread, the image is first split into square grids. The pixel with the largest gradient within each square grid is selected to be a possible feature. To ensure that each feature is unique, the pixel has to have a gradient larger than the gradient threshold before it is selected as a feature. This gradient threshold is determined by the median gradient in each $32$x$32$ pixel block. This allows the feature selection strategy to automatically find features that are unique even in low textured environments. 
This might be good for rain conditions where a large number of occlusions occur due to raindrops but good features could still be found in the regions that are not occluded. However, this design does not account for scenarios where the majority of the image is compromised and no good features could be found. Such a scenario is likely to occur in a sequence of over-exposed images, commonly seen in rain conditions. This would result in a lack of features found, leading to tracking failure.

DSO adopts outlier detection where matched points with errors above a threshold are discarded. The threshold is dependent on the median residual of the image which accounts for segments where the image is of a lower quality allowing matches with higher errors to pass the threshold. This is common in rain conditions where image quality changes randomly and would require an adaptive threshold to account for this change in image quality. When the photometric error of the current frame exceeds double the error of the previous frame, it is considered as a tracking failure and the algorithm tries to recover the localization.
This allows for the recovery of the algorithm when the quality of images improve, at the cost of losing localization for the segment of poor quality. Such a design could be good for sensor fusion methods where other sensors can support the localization task when the vision component has failed. But, it is not ideal for pure VO in rain conditions as there will be a loss in localization for that segment.

\subsection{Semi-direct Visual Odometry}
SVO \cite{forster_svo_2017} is a classical, sparse, mixed direct and indirect method that allows for both monocular and stereo setup. It was designed to strike a good balance between precision and speed for onboard computers on Micro Aerial Vehicles (MAVs), and was tested on the TUM RGB-D \cite{sturm_benchmark_2012-1}, EuroC MAV \cite{burri_euroc_2016} and ICL-NUIM \cite{handa_benchmark_2014} datasets. It achieved similar performance to DSO on the EuroC and ICL-NUIM datasets. SVO uses additional edgelet features to supplement their feature extraction strategy and uses affine illumination model to handle sudden exposure changes in the scene. 

The feature extraction strategy involves dividing the image into $32$x$32$ pixel grids and within each grid, the FAST \cite{leonardis_machine_2006} corner feature with the highest score is selected. If the grid has no corner features, the pixel with the highest gradient magnitude is selected as an edge feature. Although such a design ensures a fixed number of features are found, it might result in tracking of poor features. This allows SVO to continue tracking in segments of poor quality (over-exposure or large occlusions by raindrops) at the cost of localization accuracy. Also, it might be more susceptible to segments where only a small region of the image is distorted (small raindrop), but features in that region reduce the localization accuracy.

SVO uses a robust model for depth prediction to minimize the impact of outliers. It also uses an affine illumination model to handle the illumination change across a longer time frame. The presence of outliers and illumination change are more likely to occur in rain conditions and such a design could improve localization accuracy in rain.

\subsection{CNN-SVO}
CNN-SVO \cite{loo_cnn-svo_2018} is a mixed learning and classical method where it builds upon SVO by applying a single-image depth prediction via a convolutional neural network. This network initializes the Bayesian depth filter with a mean and variance rather than a large uncertain value range such that the filter converges to the true depth value faster, resulting in better robustness and motion estimation. As a result, CNN-SVO performs significantly better for autonomous driving applications than SVO which is designed for MAVs. CNN-SVO uses the same feature extraction and matching strategy as SVO. 

\subsection{Depth and Flow for Visual Odometry (DF-VO)}
DF-VO \cite{zhan_df-vo_2021} is a learning-based, direct monocular VO method that uses mixed dense and sparse features. It uses deep learning models for optical flow and depth prediction while using classical methods to perform pose estimation and scale recovery. DF-VO is trained using data sequences 00-08 from the KITTI dataset \cite{geiger_are_2012} and was evaluated on both the KITTI and Oxford Robotcar \cite{maddern_1_2017} datasets. DF-VO outperformed ORB-SLAM without loop closing \cite{mur-artal_orb-slam_2015} as well as DSO \cite{engel_direct_2016} and SVO \cite{forster_svo_2017} for the Oxford Robotcar Dataset. Although deep learning models were used, sparse features from the dense optical flow predictions were extracted to reduce computational cost. 

The sparse features are extracted by dividing the image into 100 regions (10x10). Then to get an even spread of features, K number of optical flow features within each region is extracted. K is determined by the number of features that pass a given threshold or the number required to extract $2000$ features in the image, whichever is lower. The threshold is determined by their proposed flow consistency metric which checks the accuracy of the predicted flow. In light rain conditions, this design helps improve localization accuracy as it removes outliers but in heavy rain conditions, it could result in tracking failure.

DF-VO also accounts for extreme scenarios where a constant velocity motion model is used to replace the tracking when insufficient features are found. This would be useful in rain conditions where it is common to find a series of over-exposed images that has minimal useful visual features.

\subsection{TartanVO}
TartanVO \cite{wang_tartanvo_2020} is a learning-based, indirect, monocular method that uses dense features. It was designed to be generalizable and without need of fine-tuning to perform well on an unseen dataset. It was trained on the TartanAir dataset \cite{wang_tartanair_2020}, an all-weather drone dataset and tested on both urban driving and aerial datasets. It performed well when compared to ORB-SLAM \cite{mur-artal_orb-slam_2015} for the KITTI dataset \cite{geiger_are_2012} as well as ORB-SLAM, DSO and SVO for the EuroC dataset \cite{burri_euroc_2016}. It does not have a feature extraction and matching strategy as it uses deep learning models for both optical flow and pose predictions. It also does not detect localization failure or outliers. This might cause it to be susceptible to extreme distortions caused by rain despite being trained on all-weather data.

\subsection{ORB-SLAM3}
ORB-SLAM3 is a classical, indirect VSLAM algorithm that uses sparse features and supports both monocular and stereo setup. It uses ORB features \cite{rublee_orb_2011} which are fast to detect and resistant to noise. ORB-SLAM3 was tested on the EuRoC \cite{burri_euroc_2016} dataset. It outperformed DSO and SVO on average for the monocular camera setup while outperforming SVO for the stereo setup. Its feature extraction and matching strategy is described in \cite{mur-artal_orb-slam_2015} where features are extracted for every frame while in contrast, DSO and SVO only extract features in the keyframes. 

ORB-SLAM3 also aims to split the features found evenly across the image where the image is divided into a grid to search for corner features. A threshold is also employed to find the best features within each cell. This threshold is reduced if insufficient features are found. Such a design ensures that sufficient features are found even for images with poor quality at the cost of identifying lower quality features. 

Outliers are detected by using an orientation consistency test and also the RANSAC procedure was used when computing the homography and fundamental matrix. Tracking also stops when insufficient correspondences are found. Thus, when the image suffers from large amounts of distortions under rain conditions, the algorithm will stop localizing and perform relocalization. 

\subsection{DROID-SLAM}
DROID-SLAM \cite{teed_droid-slam_2021} is a learning-based, indirect VSLAM method that uses dense features and supports both monocular and stereo camera setups. It was trained on the TartanAir \cite{wang_tartanair_2020} dataset, which is an all-weather synthetic drone dataset. Different from TartanVO, DROID-SLAM uses classical methods to perform bundle adjustment for pose estimation while keeping the optical flow model to perform feature matching. DROID-SLAM was evaluated on the EuroC and TartanAir datasets and outperformed both ORB-SLAM3 and TartanVO respectively. The optical flow model used for DROID-SLAM predicts both optical flow and confidence score for each of the pixel. Every pixel matching is weighted by a confidence score and used in the bundle adjustment optimization step, thereby maximizing the information used. This design ensures that tracking will not be lost even when images are compromised as the confidence value will prevent the erroneous matching from worsening the pose estimation. However, this is dependent on the accuracy of the confidence prediction by the model and would cause errors in pose estimation if a high confidence is given to a poor matching. DROID-SLAM does not detect whether tracking has failed and thus might perform poor localization in extreme conditions where no matches could be found. 

\subsection{DROID-SLAM based Heuristic Approach}
We propose a variant of DROID-SLAM algorithm to include additional map information and heuristics for the DROID-SLAM algorithm to detect and improve upon poor localization in rain conditions for stereo camera setup \cite{tan_localization_2023}. The map information could be easily obtained from any online routing services, which is used to provide a conservative global reference path (CGRP). The heuristics (H) are designed to dynamically modify the keyframe selection criteria depending on the confidence of the feature matching. The lower the confidence, the more keyframes should be taken to reduce the inaccuracies caused by a lower confidence estimation. Such a design improves localization robustness and accuracy.

\section{DATASETS}
We used three datasets from different cities as discussed below. The first dataset is Oxford Robotcar Dataset which provides multiple long-distance sequences taken along the same route at different times for an urban environment. Nine sequences were recorded in rain conditions, out of which four with sufficiently long ground truth were selected for evaluation. One sequence in clear weather along the same route was also included, where the route is roughly $9$km long. Out of these five sequences, the 2015-05-29 sequence has misaligned ground truth near the end while the 2014-11-21 sequence has an incomplete ground truth. Thus, these two sequences have a shorter route compared to the other three. The Oxford Robotcar Dataset uses the Bumblebee XB3 Trinocular stereo camera which has a $24$cm stereo baseline providing an image stream at $16$fps. The dataset provides a real-time kinematic (RTK) ground truth that is obtained by post-processing raw GPS, IMU, and static GNSS base station recordings \cite{maddern_real-time_nodate}.

The second dataset is 4Seasons Dataset which provides two rain sequences taken along different routes, namely, the 10-07 sequence from a suburban environment and the 12-22 sequence from an urban environment, both in the city of Munich. Two monochrome cameras were placed in a stereo setup with a $30$cm stereo baseline providing an image stream at $30$fps. The provided RTK-GNSS data was used as ground truth. 

An internal dataset recorded in Singapore was used for heavy rain evaluation in urban environment. Two monocular NIR cameras were set up in a stereo format with a $40$cm baseline, while the GNSS+IMU system was used as the ground truth.

\begin{figure}[h!]
    \centering
    \newcolumntype{C}{>{\centering\arraybackslash}X}
    \begin{tabularx}{\columnwidth}{CC}
    \fontfamily{cmr}\selectfont
        \includegraphics[width=0.48\columnwidth]{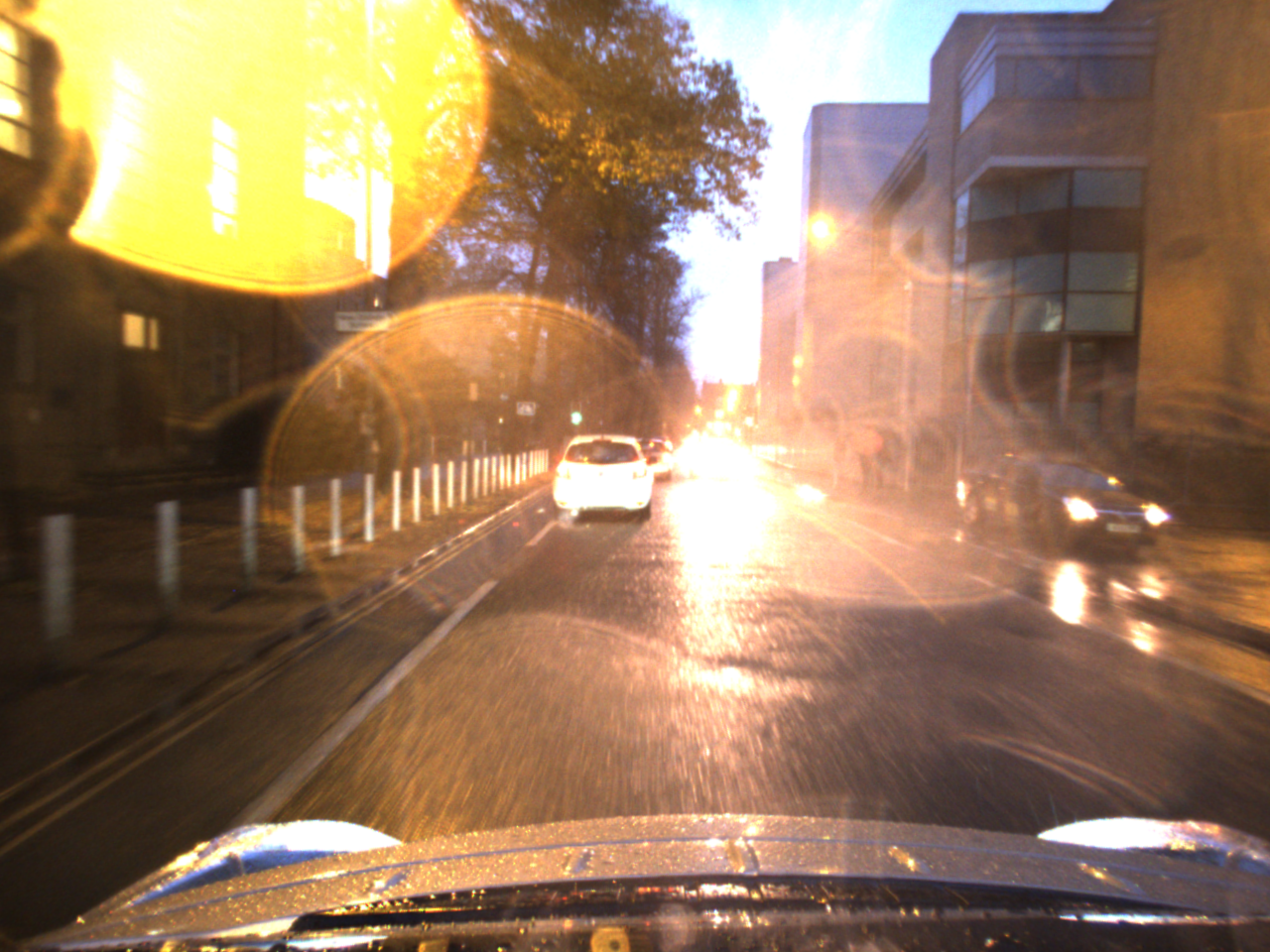} &
        \includegraphics[width=0.48\columnwidth]{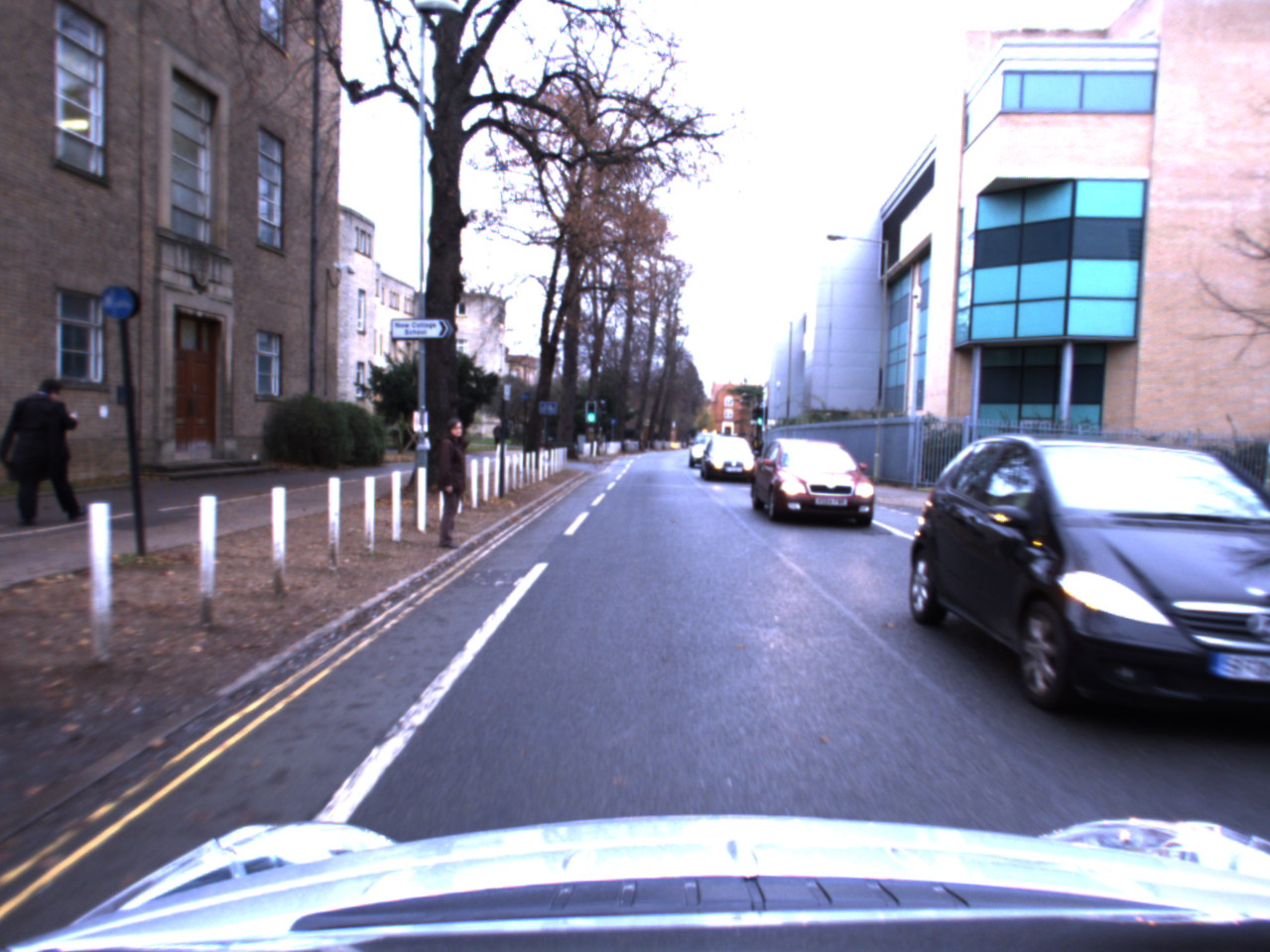}\\
        \scriptsize{$11$-$21$ (Night + Rain) Blur: \textbf{0.849}} & \scriptsize{$12$-$09$ (Clear) Blur: \textbf{0.606}} \\
    \end{tabularx}
    \caption{Comparison of blur value for the same scene taken from the $11$-$21$ and $12$-$09$ sequences respectively, from the Oxford Robotcar Dataset}
    \label{fig:compare_blur}
\vspace{-0.02\textheight}    
\end{figure}

To quantify the rain intensity from each sequence, a blur index was used as an approximation. 
Using the algorithm described in \cite{hanghang_tong_blur_2004}, the Haar wavelet transform was used to measure blurriness. We report the "BlurExtent" ratio as described in \cite{hanghang_tong_blur_2004} for each of the Tables under Section \ref{results}. Fig. \ref{fig:compare_blur} shows the blur value for the same scene taken at a different time and weather condition.

\section{EXPERIMENTAL RESULTS}
\label{section:Experiments}
\subsection{Setup}
For each algorithm evaluated, we implemented the default configuration parameters provided by the respective open-source repository for each of the datasets. The default configuration refers to the unique parameters of each method and also includes the image pre-processing steps. Only the intrinsic parameters of the cameras were modified to match the image data from each dataset. For learning-based methods, the model was also taken as it is and no fine-tuning or additional training was conducted. We note that some methods are designed for aerial context and that the default configuration might not be ideal for the urban driving context. 

For each dataset, the left image from the stereo camera setup was used for Monocular evaluation and the images were undistorted before inputting them into the VO method. The ground truth for each sequence was interpolated such that each ground truth pose corresponds to each image frame. It was also cleaned to remove any erroneous points. For the Oxford Robotcar Dataset, a later start point was chosen where the vehicle is already on the main road such that it is consistent with the start points of the other two datasets. 
In the following list, we provide algorithm specific changes that were made to run long-term localization in rainy weather.
\begin{itemize}[leftmargin=*]
\item 
DROID-SLAM is unable to run on long routes due to memory constraints as it saves every keyframe image and dense features for Global Bundle Adjustment (GBA). Thus, it was modified by removing the GBA and keyframes are forgotten after it reaches a $6$GB GPU memory capacity. This modified version of DROID-SLAM is shortened to be MDS. (MDS + CGRP + H) adds on the Conservative Global Reference Path (CGPR) and heuristics to MDS. %
    
\item 
SVO has the option to use their exposure compensation algorithm \cite{zhang_active_2017}. Given the challenging task of localization in rain, we enabled this algorithm to improve localization accuracy.

\item 
For DF-VO, we used the monodepth2 model trained on the stereo KITTI dataset.
    
\item 
For CNN-SVO, the default monodepth model is city2kitti, while we used the kitti\_resnet50 model that is trained on KITTI as it worked best for the Oxford Robotcar sequences that we used for evaluation. 
    
\item 
DSO tended to crash easily when running on the Oxford Robotcar sequences that we used for evaluation. Therefore two open source patches to the code were used in this experiment (pull request $234$ and $81$). The first pull request fixes a code implementation of the Schur complement, while the second fixes a segmentation fault bug which causes a crash when no positive IDepth is available.
    
\item 
Stereo DSO implemented uses an open source modification \cite{wu_yang_yan_li_2018} of DSO inspired by techniques used in LSD-SLAM \cite{fleet_lsd-slam_2014}. This is not to be confused with  \cite{wang_stereo_2017}. 

\item 
TartanVO was used without any changes.
\end{itemize}

\subsection{Evaluation Metric}
The Absolute Trajectory Error \cite{sturm_benchmark_2012} is used for our evaluation. The output poses are scaled and aligned (7DOF) with the ground truth for each sequence. It is also projected onto the 2D plane before evaluation.

\begin{table*}[!h]
\caption{ATE RMSE results running Monocular VO on the first 500m of all datasets (results reported in meters). ORB-SLAM3 (no lc) represents ORB-SLAM3 without loop closure. Under the $4$Seasons dataset column, $10$-$07$ represents the neighborhood\_$3$\_train sequence, $12$-$22$ represents the city\_loop\_$1$\_train sequence. Under the Oxford Robotcar dataset column, $12$-$09$ represents the $2014$-$12$-$09$-$13$-$21$-$02$ sequence, $10$-$29$ represents the $2015$-$10$-$29$-$12$-$18$-$17$ sequence, $11$-$25$ represents the $2014$-$11$-$25$-$09$-$18$-$32$ sequence, $05$-$29$ represents the $2015$-$05$-$29$-$09$-$36$-$29$ sequence, $11$-$21$ represents the $2014$-$11$-$21$-$16$-$07$-$03$ sequence.}
\begin{center}
\begin{tabular}{|c|cc|c|cccc|ccc|}
\hline
\multicolumn{ 1}{|c|}{Dataset} & \multicolumn{ 2}{c|}{4Seasons} & \multicolumn{ 5}{c|}{Oxford Robotcar Dataset} & \multicolumn{ 3}{c|}{Singapore} \\ \hline
\multicolumn{1}{|c|}{Sequences} &  12-22 & 10-07 &  12-09  &  11-25 & 05-29 &  10-29 &  11-21 & zero & one & five \\ 
 & (rain) & (rain) & (clear)  & (rain) &  (rain) & (rain) & (rain + night) & (heavy rain) & (heavy rain) & (heavy rain) \\ \hline
\multicolumn{ 1}{|c|}{Average Blur} &  0.60 & 0.63 & 0.58 & 0.62 & 0.78 & 0.79 & 0.83 & 0.80 & 0.81 & 0.88  \\ \hline
DSO & x & 67.52 & 1.19 & x & x & x & x & x* & x* & x*\\ 
SVO & 41.60 & 47.70 & x** & 31.46 & 31.41 & 15.00 & x** & x* & x* & x*\\ 
ORB-SLAM3 (no lc) & 61.69 & 20.68	& \textbf{1.12} & \textbf{1.62} & x & x & x & x & x & x\\ 
\hline
TartanVO & 68.68 & 68.07 & 24.72 & 15.21 & 18.45 & 18.89 & 13.21 & x* & 28.75 & 37.19\\ 
MDS & 57.25 &  \textbf{2.25} & 3.13 & 5.91 & 10.12 & 5.09 & 5.94 & \textbf{6.54} & 13.66 & \textbf{6.07}  \\ 
DF-VO & \textbf{6.76} & 3.06 & 3.75 & 2.98 & \textbf{5.87} & \textbf{2.73} & 7.68 & 12.23 & \textbf{3.54} & 12.02 \\ 

CNN-SVO & x* & x* & 12.31 & 21.75 & 14.49 & 12.79 & \textbf{5.08} & 18.32 & 12.58 & x*\\  \hline
\end{tabular}

\begin{tablenotes}

{\bf Types of delocalization:}
    {\bf x}: Tracking failure in the middle of the route, 
    {\bf x*}: Delayed initialization for a distance $>10m$,
    {\bf x**}: A prolonged static localization even though the vehicle is moving

\end{tablenotes}
\end{center}

\label{eval_all_mono}
\vspace{-0.01\textheight}  
\end{table*}

\begin{table*}[!h]
\caption{ATE RMSE results running Stereo VO on the Oxford Robotcar Dataset (results reported in meters). This table uses the same representations described in Table \ref{eval_all_mono}.}
\begin{center}
\begin{tabular}{|c|cc|c|cccc|ccc|}
\hline
\multicolumn{1}{|c|}{Dataset} & \multicolumn{ 2}{c|}{4Seasons} & \multicolumn{ 5}{c|}{Oxford Robotcar Dataset} & \multicolumn{ 3}{c|}{Singapore} \\ \hline
\multicolumn{ 1}{|c|}{Sequences} &  10-07 &  12-22 &  12-09 &  10-29 &  11-25  &  05-29 
&  11-21 & zero & one & five \\ 
 & (rain) & (rain) & (clear)  & (rain) &  (rain) & (rain) & (rain + night) & (heavy rain) & (heavy rain) & (heavy rain) \\ \hline
\multicolumn{ 1}{|c|}{Average Blur} & 0.62 &	0.62   & 0.61 & 0.72 & 0.74 & 0.75 & 0.80 & 0.79 & 0.84 & 0.88    \\ \hline
Stereo DSO & 159.98 & 1417.83  & x & x & x & x & x & x  & x  & x  \\ 
Stereo SVO & 152.34 & \textbf{747.35}  & 92.11 &	224.79 &	x** &	116.20 &	53.20 & x*  & x*  & x*  \\ 
ORB-SLAM3 (no lc) & x & x  & x & x & x & x & x & \textbf{21.90} & x & x \\ 
\hline
MDS & \textbf{2.11} & 1053.83 & 51.77 & 90.95 & 25.20 & \textbf{11.50} & 59.47 & 46.40 & 31.10 & 38.40 \\ 
\multicolumn{1}{|c|}{MDS + CGRP + H} & 3.38 & 786.58  & \textbf{18.49} & \textbf{14.77} & \textbf{23.18} & 11.56 & \textbf{15.62} & 44.66 & \textbf{29.31} & \textbf{32.26} \\ \hline
\end{tabular}

\begin{tablenotes}

{\bf Types of delocalization:}
    {\bf x}: Tracking failure in the middle of the route, 
    {\bf x*}: Delayed initialization for a distance $>10m$,
    {\bf x**}: A prolonged static localization even though the vehicle is moving 

\end{tablenotes}

\end{center}
\label{eval_all_stereo}

\vspace{-0.02\textheight}

\end{table*}

\section{RESULTS AND ANALYSIS} \label{results}

\subsection{Quantitative Analysis}
Monocular VO is unable to localize well for long distances and the ATE is high across all methods as well as datasets when evaluating on the entire route. We consider a vehicle to be delocalized when (a) there is a tracking failure in the middle of the route, (b) there is delayed initialization of the localization algorithm for more than $10$m and (c) there is a prolonged static localization even though the vehicle is moving. In order to make meaningful comparisons across different blur values in rain, we evaluate the first $500$m of each data sequence and reported the results in Table \ref{eval_all_mono}. The classical methods (first $3$ methods) tend to delocalize for the heavy rain sequences while the learning-based methods (last $4$ methods) were able to continue tracking. This is due to the intentional design of the classical methods which stops localization when the number of good features to track falls below a certain threshold. Such a design causes it to be less robust to challenging sequences for a pure visual approach but could be useful in a sensor fusion approach that switches between modalities.

For the classical methods, DSO and ORB-SLAM3 outperforms the learning-based methods for the 12-09 clear weather sequence. SVO delocalized for the clear weather sequence likely because of using irrelevant features selected from image regions with sky thus, resulting in erroneous localization output. However, for the heavier rain sequences, SVO is able to prevent delocalization likely due to its mixed direct and indirect model together with its affine illumination model which is used to handle exposure change and improve its robustness. For the light rain sequences from the 4Seasons dataset, all the classical methods suffer from much higher errors despite a minimal change in blurriness. This is likely due to a lack of generalizability across datasets, which displays the need for manual tuning of parameters despite all the datasets being in the same category of urban driving scenarios.

For the learning-based methods, TartanVO suffers from high localization errors throughout all data sequences due to a drifting issue described in more detail in qualitative analysis section. CNN-SVO outperforms SVO for the Oxford Robotcar and Singapore dataset, showing the usefulness of having a depth prediction model. For the 4Seasons dataset, CNN-SVO was unable to localize fully because of initialization errors. MDS performs consistently well across the data sequences except for the high error obtained on the $12$-$22$ sequence. This is caused by tracking the combined raindrop and dynamic object in the scene, which indicates that more training is required to generalize MDS to specific rain conditions. Overall, DF-VO performs consistently well across both clear and rain conditions. This robustness could be due to its outlier detection module together with its depth prediction module.

We also run the stereo camera setup for all datasets to evaluate long-term localization in both clear and rainy weather, and present its results in Table \ref{eval_all_stereo}. The classical methods delocalizes for almost all datasets. This is likely due to the build up of errors across longer data sequences. Although stereo DSO and ORB-SLAM3 are sometimes able to continue localizing after tracking failure, those data sequences are still marked as delocalized (represented as x in Table \ref{eval_all_stereo}). The MDS method suffers from high error rates in rain sequences but does perform significantly better after adding on our proposed modifications (MDS + CGRP + H) \cite{tan_localization_2023}. When comparing between monocular and stereo methods, the stereo MDS fixes the scale inconsistency problem present in the monocular MDS. 

There is no clear correlation between localization error and average blur as there are too many confounding variables such as different dynamic objects or varying traffic conditions. Future work could evaluate visual odometry with synthetic raindrops to investigate purely the effects of raindrops on the scene.

\subsection{Qualitative Analysis}
\vspace{-0.0009\textheight} 
\label{qualitative_analysis}

\begin{figure}[!h]
    \centering
    \fontfamily{cmr}\selectfont
        \includegraphics[width=.48\columnwidth]{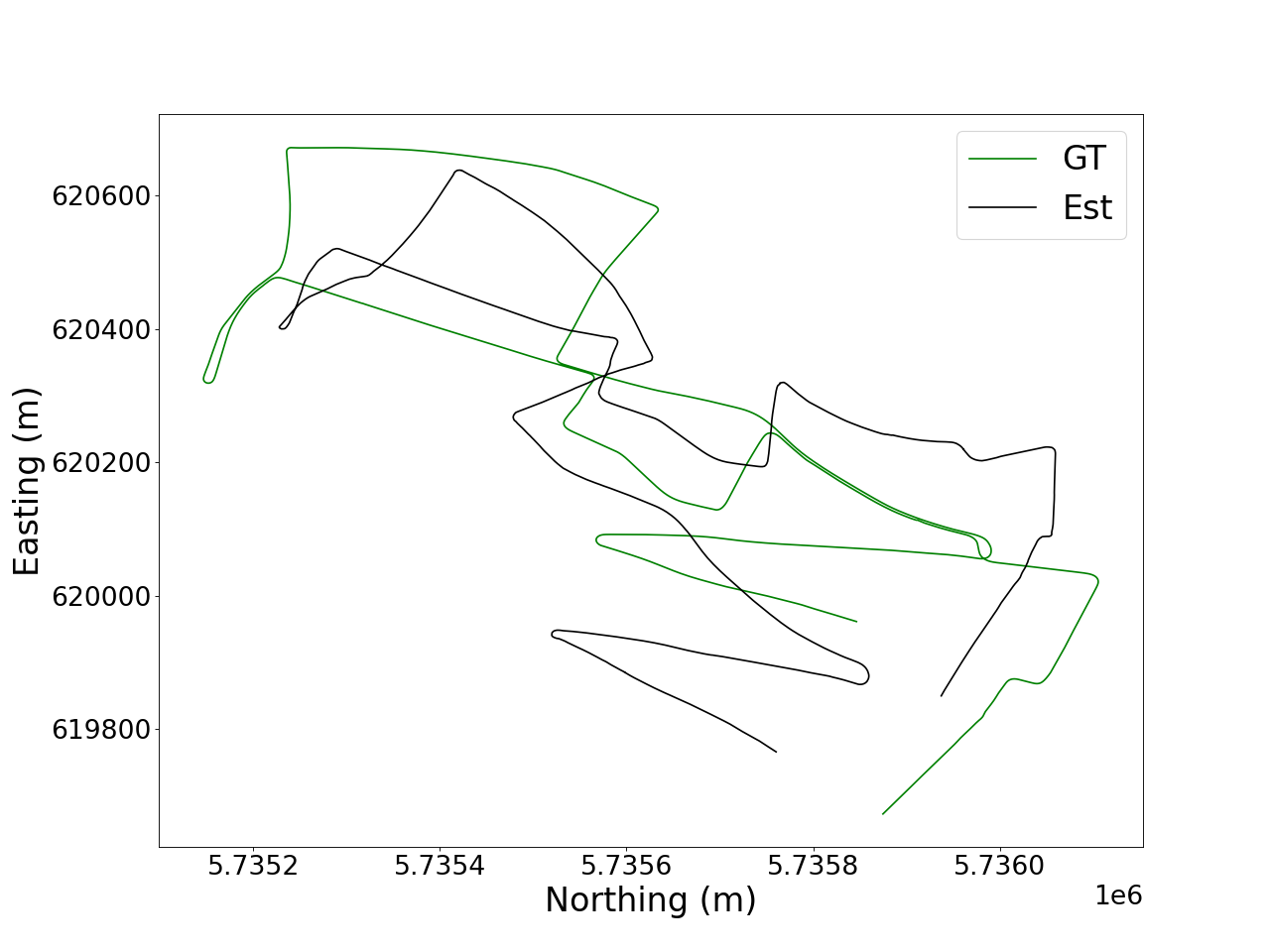}
        \includegraphics[width=.48\columnwidth]{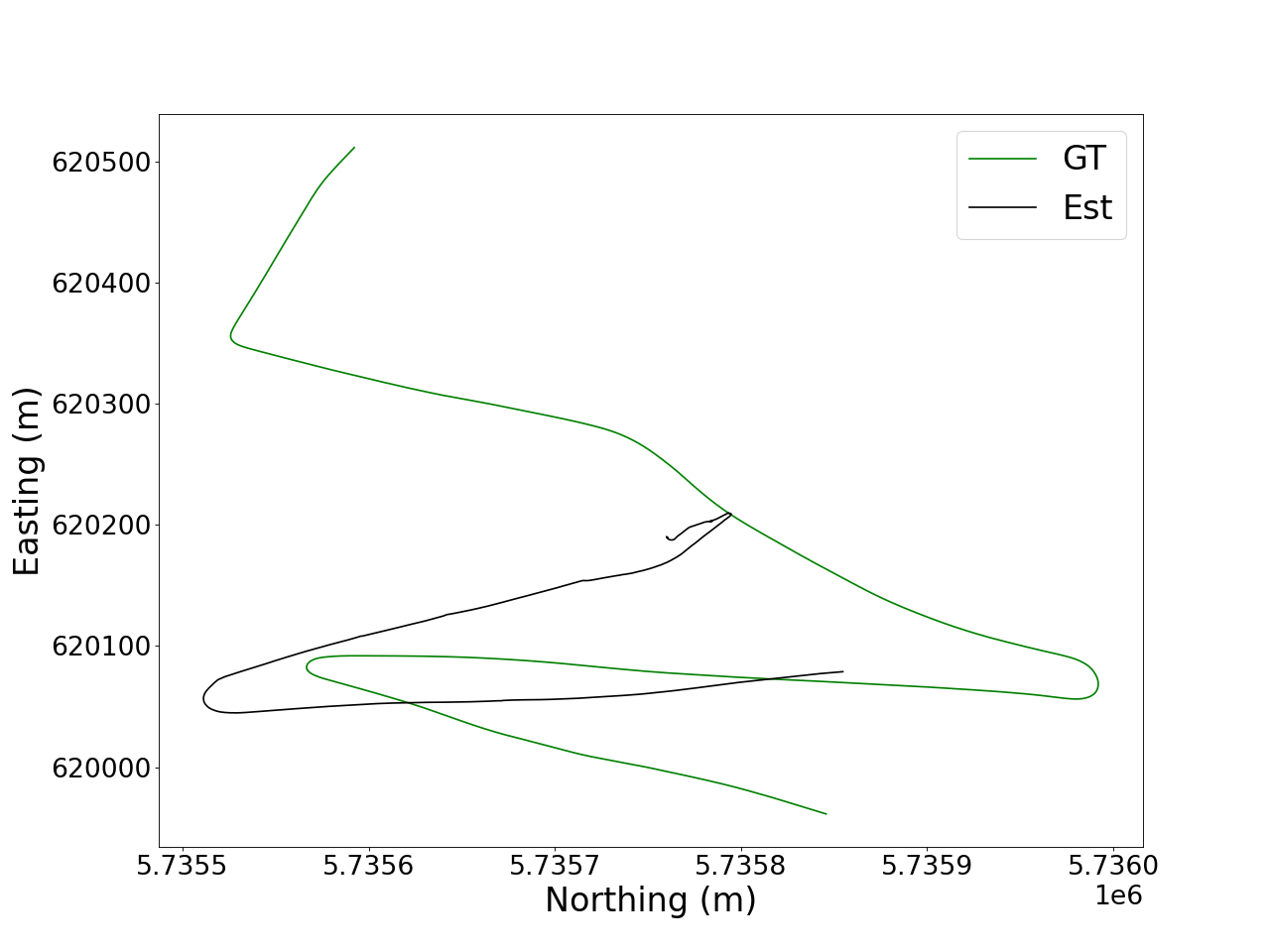} \\
        \scriptsize{DF-VO} \hspace{0.2\textwidth}
        \scriptsize{SVO} \\
    \caption{Output trajectories of DF-VO compared with SVO evaluated on the entire route of the 05-29 sequence from the Oxford Robotcar Dataset. DF-VO using a depth prediction model has a more consistent scale across the route while SVO has inconsistent scale as shown by the varying estimated path lengths compared to the ground truth.}
    \label{scale_inconsistency}
\end{figure}

\begin{figure}[!h]
    \centering
    \fontfamily{cmr}\selectfont
        \includegraphics[width=0.5\columnwidth]{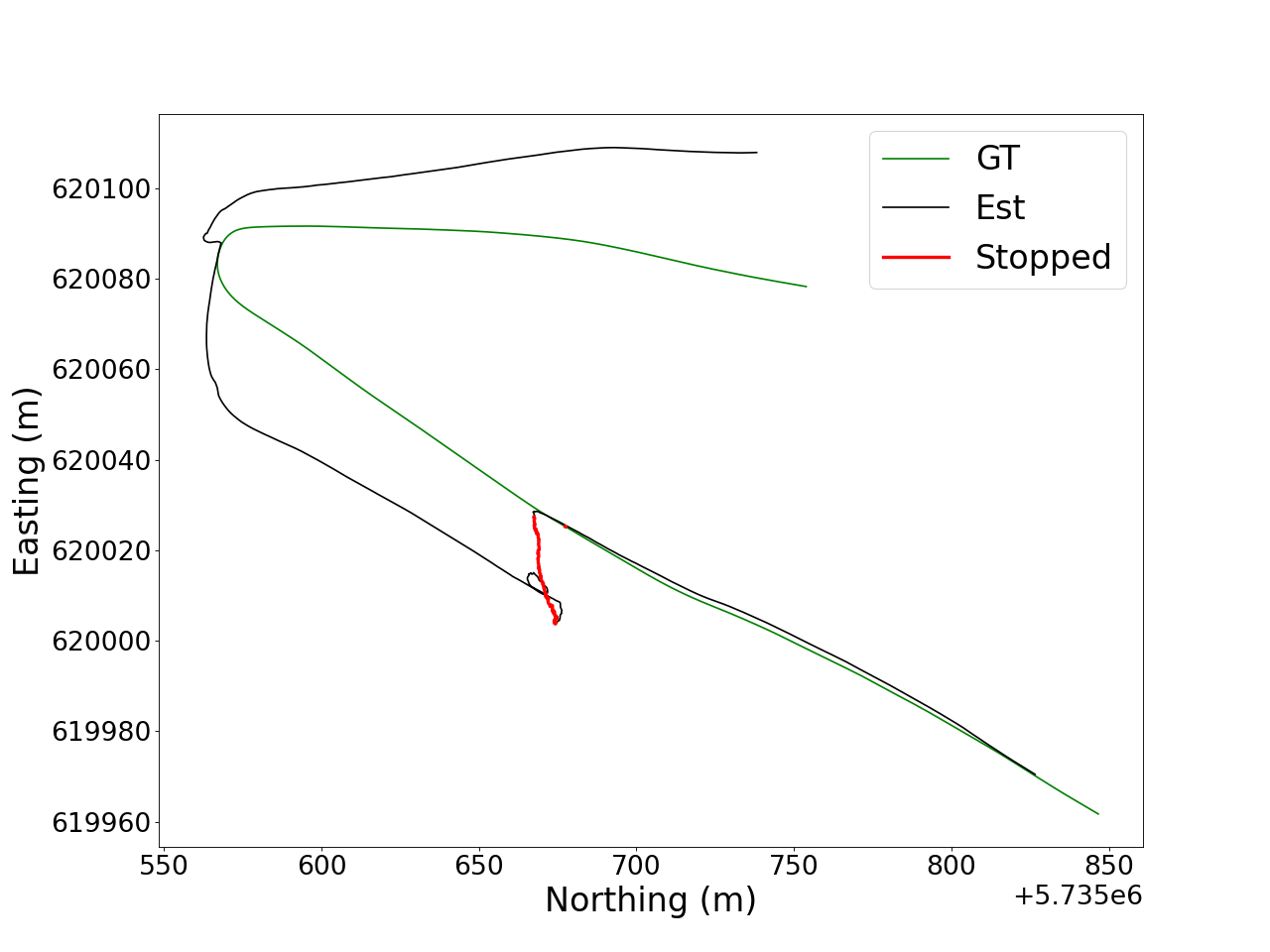} \\
        \scriptsize{First 500m of 10-29} \\
    \caption{TartanVO evaluation output on the first 500m of the 10-29 sequence from the Oxford Robotcar dataset. The red line shows the segment of the route where the vehicle is not moving in the video.}
    \label{tartan_drift}
\end{figure}

The methods such as DF-VO and CNN-SVO which employ a depth prediction model are able to maintain a consistent scale even for rain sequences, while other methods suffer from scale inconsistency problems shown in Fig. \ref{scale_inconsistency}. TartanVO suffers from drifting issues when the vehicle is at rest, as shown in Fig. \ref{tartan_drift}. This might be due to training bias from the constant motion experienced by an aerial vehicle even when hovering in place. DSO, SVO and ORB-SLAM3 delocalize easily when encountering large exposure change, likely due to the lack of matching features. SVO is more resistant to such changes due to the exposure compensation algorithm and thus is able to localize for more rain sequences.
The first 500m of the 12-22 sequence is particularly challenging due to large dynamic objects, which explains the high errors seen in Table \ref{eval_all_mono}. For monocular camera setup, DF-VO is able to identify the outliers caused by the dynamic objects which reduces the errors significantly. 

For the entire route of 12-22 sequence in the 4Seasons dataset, the stereo camera setup fails due to a challenging tunnel segment. The best localization results in the stereo camera setup were obtained from our proposed variant of modified DROID-SLAM (MDS + CGRP + H) algorithm.

In general, the occlusion caused by the adherent raindrop does not significantly impact visual odometry as long as sufficient good visual features can still be found in the scene. This is difficult for the rain + night sequence as the unoccluded regions of the images does not provide good visual features due to the night time imaging. The lens flare and rain streaks might result in undesirable visual features identified which calls for additional filters to ignore or remove this effect. Additional issues such as over-exposure when making a turn could cause the visual odometry to lose all visual features for that short period of time which calls for the need of a sensor fusion approach.

\section{CONCLUSION}
An evaluation of a wide range of VO methods was done for both clear and rain datasets. We found that the VO methods that employ a depth prediction model are able to maintain a consistent scale even for rain sequences. The stereo setup is also able to provide scale information but would require additional map information to perform well for long-term localization. Classical methods tend to delocalize easily in rain and is not recommended for rain condition unless paired with other sensors for a sensor fusion approach. Out of all the monocular methods, not one method performed the best across all three datasets. However, DF-VO performed consistently well out of all the evaluated approaches and could be adopted for a sensor fusion approach for short localization in rain conditions. For longer localization sequences the stereo method could be considered, where our proposed approach (MDS + CGRP + H) performs the best out of all the evaluated approaches across the three datasets. In conclusion, all evaluated VO approaches are insufficient to localize in rain. Hence, a more robust sensor fusion based approach is required for autonomous urban driving in rain.

\section*{Acknowledgment}
This research is supported by the Ministry of Education, Singapore, under its Academic Research Fund Tier 2 MOE-T2EP50121-0022.

\bibliographystyle{unsrt}
\bibliography{CASE_reference_list2}

\begin{thebibliography}{10}

\bibitem{garg_vision_2007}
Kshitiz Garg and Shree~K. Nayar.
\newblock Vision and {Rain}.
\newblock {\em International Journal of Computer Vision}, 75(1):3--27, July
  2007.

\bibitem{ulfwi_adherent_nodate}
Björn Ulfwi.
\newblock Adherent {Raindrop} {Detection}.

\bibitem{bahnsen_rain_2019}
Chris~H. Bahnsen and Thomas~B. Moeslund.
\newblock Rain {Removal} in {Traffic} {Surveillance}: {Does} it {Matter}?
\newblock {\em IEEE Transactions on Intelligent Transportation Systems},
  20(8):2802--2819, August 2019.

\bibitem{tan_localization_2023}
Yu~Xiang Tan, Malika Meghjani, and Marcel~Bartholomeus Prasetyo.
\newblock Localization with {Anticipation} for {Autonomous} {Urban} {Driving}
  in {Rain}, June 2023.
\newblock arXiv:2306.09134 [cs].

\bibitem{maddern_1_2017}
Will Maddern, Geoffrey Pascoe, Chris Linegar, and Paul Newman.
\newblock 1 year, 1000 km: {The} {Oxford} {RobotCar} dataset.
\newblock {\em The International Journal of Robotics Research}, 36(1):3--15,
  January 2017.

\bibitem{wenzel_4seasons_2020}
Patrick Wenzel, Rui Wang, Nan Yang, Qing Cheng, Qadeer Khan, Lukas von
  Stumberg, Niclas Zeller, and Daniel Cremers.
\newblock {4Seasons}: {A} {Cross}-{Season} {Dataset} for {Multi}-{Weather}
  {SLAM} in {Autonomous} {Driving}, October 2020.
\newblock arXiv:2009.06364 [cs].

\bibitem{zhang_autonomous_2021}
Yuxiao Zhang, Alexander Carballo, Hanting Yang, and Kazuya Takeda.
\newblock Autonomous {Driving} in {Adverse} {Weather} {Conditions}: {A}
  {Survey}, December 2021.
\newblock arXiv:2112.08936 [cs].

\bibitem{hong_radarslam_2020}
Ziyang Hong, Yvan Petillot, and Sen Wang.
\newblock {RadarSLAM}: {Radar} based {Large}-{Scale} {SLAM} in {All}
  {Weathers}.
\newblock In {\em 2020 {IEEE}/{RSJ} {International} {Conference} on
  {Intelligent} {Robots} and {Systems} ({IROS})}, pages 5164--5170, October
  2020.
\newblock ISSN: 2153-0866.

\bibitem{ort_autonomous_2020}
Teddy Ort, Igor Gilitschenski, and Daniela Rus.
\newblock Autonomous {Navigation} in {Inclement} {Weather} {Based} on a
  {Localizing} {Ground} {Penetrating} {Radar}.
\newblock {\em IEEE Robotics and Automation Letters}, 5(2):3267--3274, April
  2020.
\newblock Conference Name: IEEE Robotics and Automation Letters.

\bibitem{barnes_under_2020}
Dan Barnes and Ingmar Posner.
\newblock Under the {Radar}: {Learning} to {Predict} {Robust} {Keypoints} for
  {Odometry} {Estimation} and {Metric} {Localisation} in {Radar}.
\newblock In {\em 2020 {IEEE} {International} {Conference} on {Robotics} and
  {Automation} ({ICRA})}, pages 9484--9490, May 2020.
\newblock ISSN: 2577-087X.

\bibitem{campos_orb-slam3_2021}
Carlos Campos, Richard Elvira, Juan J.~Gómez Rodríguez, José M.~M. Montiel,
  and Juan~D. Tardós.
\newblock {ORB}-{SLAM3}: {An} {Accurate} {Open}-{Source} {Library} for
  {Visual}, {Visual}-{Inertial} and {Multi}-{Map} {SLAM}.
\newblock {\em IEEE Transactions on Robotics}, 37(6):1874--1890, December 2021.
\newblock arXiv: 2007.11898.

\bibitem{von_stumberg_direct_2018}
Lukas von Stumberg, Vladyslav Usenko, and Daniel Cremers.
\newblock Direct {Sparse} {Visual}-{Inertial} {Odometry} using {Dynamic}
  {Marginalization}.
\newblock In {\em 2018 {IEEE} {International} {Conference} on {Robotics} and
  {Automation} ({ICRA})}, pages 2510--2517, May 2018.
\newblock arXiv:1804.05625 [cs].

\bibitem{kiss-illes_gps-slam_2019}
Dániel Kiss-Illés, Cristina Barrado, and Esther Salamí.
\newblock {GPS}-{SLAM}: {An} {Augmentation} of the {ORB}-{SLAM} {Algorithm}.
\newblock {\em Sensors (Basel, Switzerland)}, 19(22):4973, November 2019.

\bibitem{brubaker_map-based_2016}
Marcus~A. Brubaker, Andreas Geiger, and Raquel Urtasun.
\newblock Map-{Based} {Probabilistic} {Visual} {Self}-{Localization}.
\newblock {\em IEEE Transactions on Pattern Analysis and Machine Intelligence},
  38(4):652--665, April 2016.

\bibitem{revaud_r2d2_2019}
Jerome Revaud, Philippe Weinzaepfel, César De~Souza, Noe Pion, Gabriela
  Csurka, Yohann Cabon, and Martin Humenberger.
\newblock {R2D2}: {Repeatable} and {Reliable} {Detector} and {Descriptor}.
\newblock {\em arXiv:1906.06195 [cs]}, June 2019.
\newblock arXiv: 1906.06195.

\bibitem{dusmanu_d2-net_2019}
Mihai Dusmanu, Ignacio Rocco, Tomas Pajdla, Marc Pollefeys, Josef Sivic,
  Akihiko Torii, and Torsten Sattler.
\newblock D2-{Net}: {A} {Trainable} {CNN} for {Joint} {Description} and
  {Detection} of {Local} {Features}.
\newblock In {\em 2019 {IEEE}/{CVF} {Conference} on {Computer} {Vision} and
  {Pattern} {Recognition} ({CVPR})}, pages 8084--8093, Long Beach, CA, USA,
  June 2019. IEEE.

\bibitem{yu_visual_2021}
Huaiyuan Yu, Haijiang Zhu, and Fengrong Huang.
\newblock Visual {Simultaneous} {Localization} and {Mapping} ({SLAM}) {Based}
  on {Blurred} {Image} {Detection}.
\newblock {\em Journal of Intelligent \& Robotic Systems}, 103(1):12, September
  2021.

\bibitem{porav_adversarial_2018}
Horia Porav, Will Maddern, and Paul Newman.
\newblock Adversarial {Training} for {Adverse} {Conditions}: {Robust} {Metric}
  {Localisation} {Using} {Appearance} {Transfer}.
\newblock In {\em 2018 {IEEE} {International} {Conference} on {Robotics} and
  {Automation} ({ICRA})}, pages 1011--1018, May 2018.
\newblock ISSN: 2577-087X.

\bibitem{yousif_overview_2015}
Khalid Yousif, Alireza Bab-Hadiashar, and Reza Hoseinnezhad.
\newblock An {Overview} to {Visual} {Odometry} and {Visual} {SLAM}:
  {Applications} to {Mobile} {Robotics}.
\newblock {\em Intelligent Industrial Systems}, 1(4):289--311, December 2015.

\bibitem{abaspur_kazerouni_survey_2022}
Iman Abaspur~Kazerouni, Luke Fitzgerald, Gerard Dooly, and Daniel Toal.
\newblock A survey of state-of-the-art on visual {SLAM}.
\newblock {\em Expert Systems with Applications}, 205:117734, November 2022.

\bibitem{agostinho_practical_2022}
Lucas~R. Agostinho, Nuno~M. Ricardo, Maria~I. Pereira, Antoine Hiolle, and
  Andry~M. Pinto.
\newblock A {Practical} {Survey} on {Visual} {Odometry} for {Autonomous}
  {Driving} in {Challenging} {Scenarios} and {Conditions}.
\newblock {\em IEEE Access}, 10:72182--72205, 2022.

\bibitem{engel_direct_2016}
Jakob Engel, Vladlen Koltun, and Daniel Cremers.
\newblock Direct {Sparse} {Odometry}, October 2016.
\newblock arXiv:1607.02565 [cs].

\bibitem{zhan_df-vo_2021}
Huangying Zhan, Chamara~Saroj Weerasekera, Jia-Wang Bian, Ravi Garg, and Ian
  Reid.
\newblock {DF}-{VO}: {What} {Should} {Be} {Learnt} for {Visual} {Odometry}?,
  March 2021.
\newblock arXiv:2103.00933 [cs].

\bibitem{fleet_lsd-slam_2014}
Jakob Engel, Thomas Schöps, and Daniel Cremers.
\newblock {LSD}-{SLAM}: {Large}-{Scale} {Direct} {Monocular} {SLAM}.
\newblock In David Fleet, Tomas Pajdla, Bernt Schiele, and Tinne Tuytelaars,
  editors, {\em Computer {Vision} – {ECCV} 2014}, volume 8690, pages
  834--849. Springer International Publishing, Cham, 2014.
\newblock Series Title: Lecture Notes in Computer Science.

\bibitem{teed_droid-slam_2021}
Zachary Teed and Jia Deng.
\newblock {DROID}-{SLAM}: {Deep} {Visual} {SLAM} for {Monocular}, {Stereo}, and
  {RGB}-{D} {Cameras}.
\newblock In {\em Advances in {Neural} {Information} {Processing} {Systems}},
  volume~34, pages 16558--16569. Curran Associates, Inc., 2021.

\bibitem{wang_tartanvo_2020}
Wenshan Wang, Yaoyu Hu, and Sebastian Scherer.
\newblock {TartanVO}: {A} {Generalizable} {Learning}-based {VO}, October 2020.
\newblock arXiv:2011.00359 [cs].

\bibitem{schenk_robust_2017}
Fabian Schenk and Friedrich Fraundorfer.
\newblock Robust edge-based visual odometry using machine-learned edges.
\newblock In {\em 2017 {IEEE}/{RSJ} {International} {Conference} on
  {Intelligent} {Robots} and {Systems} ({IROS})}, pages 1297--1304, September
  2017.
\newblock ISSN: 2153-0866.

\bibitem{gomez-ojeda_robust_2016}
Ruben Gomez-Ojeda and Javier Gonzalez-Jimenez.
\newblock Robust stereo visual odometry through a probabilistic combination of
  points and line segments.
\newblock In {\em 2016 {IEEE} {International} {Conference} on {Robotics} and
  {Automation} ({ICRA})}, pages 2521--2526, May 2016.

\bibitem{forster_svo_2014}
Christian Forster, Matia Pizzoli, and Davide Scaramuzza.
\newblock {SVO}: {Fast} semi-direct monocular visual odometry.
\newblock In {\em 2014 {IEEE} {International} {Conference} on {Robotics} and
  {Automation} ({ICRA})}, pages 15--22, May 2014.
\newblock ISSN: 1050-4729.

\bibitem{loo_cnn-svo_2018}
Shing~Yan Loo, Ali~Jahani Amiri, Syamsiah Mashohor, Sai~Hong Tang, and Hong
  Zhang.
\newblock {CNN}-{SVO}: {Improving} the {Mapping} in {Semi}-{Direct} {Visual}
  {Odometry} {Using} {Single}-{Image} {Depth} {Prediction}, October 2018.
\newblock arXiv:1810.01011 [cs].

\bibitem{burri_euroc_2016}
Michael Burri, Janosch Nikolic, Pascal Gohl, Thomas Schneider, Joern Rehder,
  Sammy Omari, Markus~W Achtelik, and Roland Siegwart.
\newblock The {EuRoC} micro aerial vehicle datasets.
\newblock {\em The International Journal of Robotics Research},
  35(10):1157--1163, September 2016.

\bibitem{handa_benchmark_2014}
Ankur Handa, Thomas Whelan, John McDonald, and Andrew~J. Davison.
\newblock A benchmark for {RGB}-{D} visual odometry, {3D} reconstruction and
  {SLAM}.
\newblock In {\em 2014 {IEEE} {International} {Conference} on {Robotics} and
  {Automation} ({ICRA})}, pages 1524--1531, May 2014.
\newblock ISSN: 1050-4729.

\bibitem{forster_svo_2017}
Christian Forster, Zichao Zhang, Michael Gassner, Manuel Werlberger, and Davide
  Scaramuzza.
\newblock {SVO}: {Semidirect} {Visual} {Odometry} for {Monocular} and
  {Multicamera} {Systems}.
\newblock {\em IEEE Transactions on Robotics}, 33(2):249--265, April 2017.
\newblock Conference Name: IEEE Transactions on Robotics.

\bibitem{sturm_benchmark_2012-1}
Jürgen Sturm, Nikolas Engelhard, Felix Endres, Wolfram Burgard, and Daniel
  Cremers.
\newblock A benchmark for the evaluation of {RGB}-{D} {SLAM} systems.
\newblock In {\em 2012 {IEEE}/{RSJ} {International} {Conference} on
  {Intelligent} {Robots} and {Systems}}, pages 573--580, October 2012.
\newblock ISSN: 2153-0866.

\bibitem{leonardis_machine_2006}
Edward Rosten and Tom Drummond.
\newblock Machine {Learning} for {High}-{Speed} {Corner} {Detection}.
\newblock In Aleš Leonardis, Horst Bischof, and Axel Pinz, editors, {\em
  Computer {Vision} – {ECCV} 2006}, volume 3951, pages 430--443. Springer
  Berlin Heidelberg, Berlin, Heidelberg, 2006.
\newblock Series Title: Lecture Notes in Computer Science.

\bibitem{geiger_are_2012}
A.~Geiger, P.~Lenz, and R.~Urtasun.
\newblock Are we ready for autonomous driving? {The} {KITTI} vision benchmark
  suite.
\newblock In {\em 2012 {IEEE} {Conference} on {Computer} {Vision} and {Pattern}
  {Recognition}}, pages 3354--3361, Providence, RI, June 2012. IEEE.

\bibitem{mur-artal_orb-slam_2015}
Raul Mur-Artal, J.~M.~M. Montiel, and Juan~D. Tardos.
\newblock {ORB}-{SLAM}: a {Versatile} and {Accurate} {Monocular} {SLAM}
  {System}.
\newblock {\em IEEE Transactions on Robotics}, 31(5):1147--1163, October 2015.
\newblock arXiv: 1502.00956.

\bibitem{wang_tartanair_2020}
Wenshan Wang, Delong Zhu, Xiangwei Wang, Yaoyu Hu, Yuheng Qiu, Chen Wang, Yafei
  Hu, Ashish Kapoor, and Sebastian Scherer.
\newblock {TartanAir}: {A} {Dataset} to {Push} the {Limits} of {Visual} {SLAM},
  August 2020.
\newblock arXiv:2003.14338 [cs].

\bibitem{rublee_orb_2011}
Ethan Rublee, Vincent Rabaud, Kurt Konolige, and Gary Bradski.
\newblock {ORB}: {An} efficient alternative to {SIFT} or {SURF}.
\newblock In {\em 2011 {International} {Conference} on {Computer} {Vision}},
  pages 2564--2571, November 2011.
\newblock ISSN: 2380-7504.

\bibitem{maddern_real-time_nodate}
Will Maddern, Geoffrey Pascoe, Matthew Gadd, Dan Barnes, Brian Yeomans, and
  Paul Newman.
\newblock Real-time {Kinematic} {Ground} {Truth} for the {Oxford} {RobotCar}
  {Dataset}.
\newblock page~3.

\bibitem{hanghang_tong_blur_2004}
{Hanghang Tong}, {Mingjing Li}, {Hongjiang Zhang}, and {Changshui Zhang}.
\newblock Blur detection for digital images using wavelet transform.
\newblock In {\em 2004 {IEEE} {International} {Conference} on {Multimedia} and
  {Expo} ({ICME}) ({IEEE} {Cat}. {No}.{04TH8763})}, pages 17--20, Taipei,
  Taiwan, 2004. IEEE.

\bibitem{zhang_active_2017}
Zichao Zhang, Christian Forster, and Davide Scaramuzza.
\newblock Active exposure control for robust visual odometry in {HDR}
  environments.
\newblock In {\em 2017 {IEEE} {International} {Conference} on {Robotics} and
  {Automation} ({ICRA})}, pages 3894--3901, May 2017.

\bibitem{wu_yang_yan_li_2018}
Jiatian Wu, Degang Yang, Qinrui Yan, and Shixin Li.
\newblock Stereo-dso.
\newblock url={https://github.com/JiatianWu/stereo-dso}, 2018.

\bibitem{wang_stereo_2017}
Rui Wang, Martin Schwörer, and Daniel Cremers.
\newblock Stereo {DSO}: {Large}-{Scale} {Direct} {Sparse} {Visual} {Odometry}
  with {Stereo} {Cameras}, August 2017.
\newblock arXiv:1708.07878 [cs].

\bibitem{sturm_benchmark_2012}
Jürgen Sturm, Nikolas Engelhard, Felix Endres, Wolfram Burgard, and Daniel
  Cremers.
\newblock A benchmark for the evaluation of {RGB}-{D} {SLAM} systems.
\newblock In {\em 2012 {IEEE}/{RSJ} {International} {Conference} on
  {Intelligent} {Robots} and {Systems}}, pages 573--580, October 2012.
\newblock ISSN: 2153-0866.

\end{thebibliography}

\end{document}